\documentclass[french,utf8x]{article}

\usepackage{geometry}
\usepackage[export]{adjustbox}
\usepackage{url}
\usepackage{tipa}
\usepackage{babel}

\newcommand{\AZee}[1]{\texttt{#1}}

\begin{document}

\title{Représentation graphique de la langue des signes française et édition logicielle}
\maketitle

\begin{center}
\author{Michael Filhol\fup{1,2} \& Thomas von Ascheberg\fup{1} \\
\fup{1} LISN, CNRS\\
\fup{2} Université Paris--Saclay
}
\end{center}

\abstract{
Cet article propose une méthode pour définir une forme graphique éditable standardisée pour les langues des signes, ainsi qu'une proposition "AZVD" et un éditeur logiciel associé. Inspirée d'une part par les régularités observées dans les pratiques spontanées de locuteurs pratiquant la schématisation, la démarche tente garantir un système qualifié d'adoptable. Liée d'autre part au modèle formel de représentation AZee, elle vise également à spécifier un système dont toutes les productions ont une lecture déterminée au point où elles sont automatiquement synthétisables par un avatar.
}


\section{Introduction}
Les langues dotées d'une forme écrite sont en général facilement équipées de logiciels d'édition voire de traitements automatiques en conséquence.
Ce n'est en revanche pas le cas des langues des signes (LS) qui en sont dépourvues.
La vidéo est parfois utilisée comme medium de substitution, mais sa manipulation est souvent lourde et fastidieuse : son stockage est démesurément volumineux, son édition est impossible au sens de l'équivalent d'un traitement de texte, sa requêtabilité et son indexation sont extrêmement limitées...
Aussi elle dépend fortement d'un accès à la technologie et, comparable plutôt à un enregistrement vocal qu'une représentation écrite, reste liée au temps réel de la production tant pour enregistrer que visionner, alors qu'une écriture permet la prise de note avec papier et crayon, lisible indépendamment du temps réel.
Pis encore, son anonymat est toujours compromis, ce qui a des conséquences non négligeables sur la diffusion des informations et des idées, par exemple sur internet.

Ainsi cet article propose-t-il de réfléchir à la standardisation d'une représentation éditable pour la langue des signes française (LSF) pouvant être intégrée et manipulée dans des logiciels tels que, pour commencer, des éditeurs.
Pour ce faire, nous présentons d'abord quelques systèmes de représentation conçus pour les LS d'une part, que nous contrastons avec une pratique de schématisation graphique des LS que nous présentons aussi, plus en détail.
Celle-ci faisant émerger des régularités spontanées, nous présentons ensuite une manière de les rapprocher d'AZee, un modèle de représentation formel des énoncés LS que nous présentons en résumé aussi, afin de créer la proposition graphique AZVD.
Une section présente enfin l'éditeur que nous avons développé pour mettre la proposition en pratique, avant de conclure avec des perspectives d'utilisation, d'évolution et de discussion.

\section{Représentations des LS, systématiques et spontanées}

\begin{figure}
    \centering
    \begin{tabular}{ccc}
        \includegraphics[height=4ex]{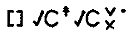}
         & \includegraphics[height=15mm]{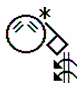}
         & \includegraphics[height=4ex]{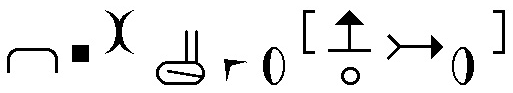}
        \\
        Notation de Stokoe
         & SignWriting
         & HamNoSys
    \end{tabular}
    \caption{Exemples de représentations graphiques conçues pour les LS}
    \label{fig:repr-conçues}
\end{figure}

Des systèmes de représentation ont été conçus pour les LS, avec des objectifs variés.
On peut distinguer les systèmes de notation destinés à la description théorique, autrement dit aux linguistes ou aux spécialistes, ceux destinés aux usagers de la langue et se positionnant comme des systèmes d'écriture, et ceux destinés aux applications informatiques.
La figure~\ref{fig:repr-conçues} en illustre trois, à savoir la notation de Stokoe \cite{stokoe65-dict}, SignWriting \cite{sutton14-textBook} et HamNoSys \cite{pri89,Hanke04-HamNoSys}.
S'ils comportent des différences dans leur apparence, leurs unités élémentaires ou la manière de les combiner, ces modèles réduisent tous les énoncés signés à une suite d'unités dites lexicales dont on donne les paramètres manuels \cite{sto60} ainsi que, pour certains, des éléments non-manuels qui les accompagnent.

Tous ces scripts sont conçus et envisagés comme des \emph{systèmes} avec la vocation de déterminer leur lecture (formes gestuelles à articuler), avec une cohérence interne et une régularité au sens où à un même énoncé correspond une même représentation, et vice-versa.
Ceci les rapprocherait des systèmes d'écriture connus pour les langues vocales, non signées.
On note cependant qu'aucun n'a été adopté au sens large par les différentes communautés de signeurs, alors même que certains ont bénéficié d'efforts de promotion et d'outils pour leur diffusion.
En particulier, on peut citer plusieurs ressources\footnote{La principale est le \emph{SignPuddle} en ligne : \url{https://www.signbank.org/signpuddle}.} et éditeurs logiciels\footnote{Plusieurs sont répertoriés ici : \url{https://www.signwriting.org/downloads}.} pour SignWriting, ou des programmes de synthèse par signeur virtuel, par exemple pour HamNoSys \cite{ell08}.
Leur manque d'adoption contraste donc avec les systèmes d'écriture adoptés pour les langues écrites non signées.

\begin{figure}
    \centering
    \includegraphics[width=.75\columnwidth]{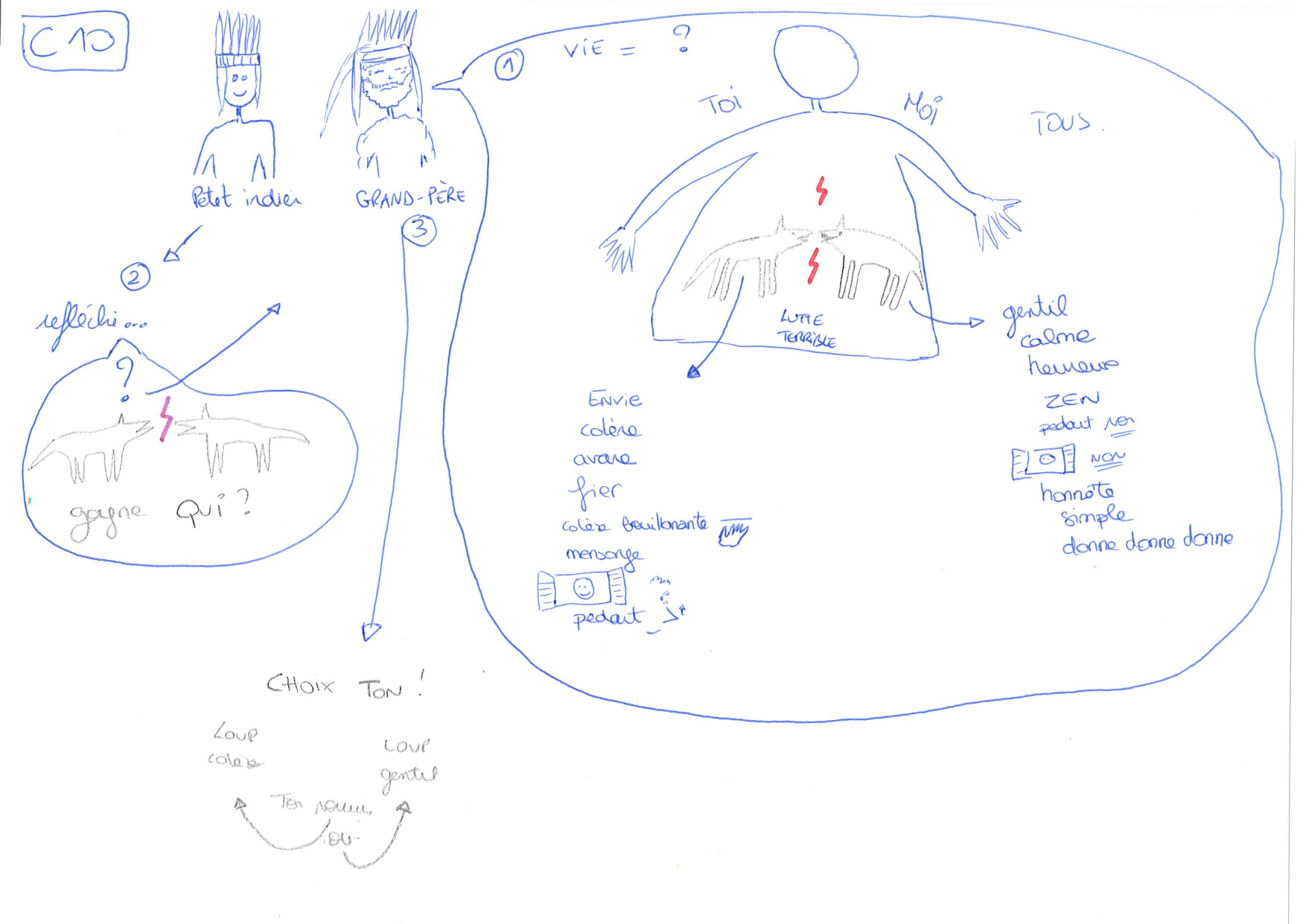}
    \caption{Schématisation d'une narration en LSF}
    \label{fig:VD-complet}
\end{figure}

Ce manque est d'autant plus notable que le besoin d'une représentation semble pourtant exister.
En effet, les sourds veulent parfois prendre des notes ou préparer un discours en se donnant un prompteur pour une lecture sans coupe devant une caméra par exemple.
Ou encore, les traducteurs sont souvent formés à travailler leur résultat en langue cible de manière détachée de la forme source.
Pour ce faire, les locuteurs ont alors recours à des schématisations sur papier, mettant en jeu des icônes et dessins, des lignes et flèches pour les relier, etc.
La figure~\ref{fig:VD-complet} donne un exemple de représentation d'histoire courte dont la production signée dure environ une minute.

Ces productions, que nous appelons ``schémas de verbalisation'' (VD pour \emph{verbalising diagram}), sont spontanées, en tout point librement composées par leur auteur plutôt que conformes à des règles apprises ou posées a priori.
Elles ne servent d'ailleurs en général qu'à une utilisation personnelle, souvent même unique avant d'être jetées.
Les éléments du contenu ne déterminent pas leur lecture de manière systématique.
Ceci rend difficile l'outillage informatique de la pratique, et impossible sa synthèse automatique par un avatar en particulier.
Elles contrastent donc en cela avec les systèmes conçus présentés plus haut, qu'ils soient pour les LS ou non.

Cela dit, des régularités ont été observées dans les VD, aussi bien intra- qu'inter-personnelles, comme rapporté suite à une première collecte contrôlée d'un corpus de productions VD spontanées et à son étude \cite{LREC-SL-2020-VD}.
Une première tendance nettement identifiée est celle pour les symboles de représenter le \emph{sens} de ce qu'ils désignent plutôt que la \emph{forme} à réaliser à la lecture.
Par exemple, aucun scripteur n'a représenté le soleil avec une référence à la position (près de la tempe) ou la forme (en sphère) de la main active pour produire le signe correspondant.
Tous ont procédé à un dessin de soleil ou, plus rarement, le mot écrit emprunté au français.

Ces options ne sont pas exclusives, l'iconicité de la LS permettant parfois de rapprocher un symbole à la fois de sens et forme, mais la disproportion reste grande, et le choix de la représentation du sens majoritaire.
Les schémas contrastent donc, en cela aussi, avec les systèmes conçus pour les LS.
Au contraire, le mélange fortement déséquilibré de phonographie (représentation des phonèmes -- forme articulée) et de logographie (représentation par morphèmes -- sens) dans les scripts des langues vocales écrites est en revanche la norme.
Ceci pourrait expliquer la non-adoption des systèmes conçus, la totalité d'entre eux définissant des sytèmes entièrement phonographiques.

\begin{figure}
    \centering
    \begin{tabular}{ccc}
        Symboles VD observés
         & Sens associé
         & Signe correspondant
        \\ \hline
        \includegraphics[height=6ex]{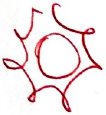}\hspace{5mm}
        \includegraphics[height=6ex]{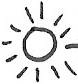}\hspace{5mm}
        \includegraphics[height=4ex]{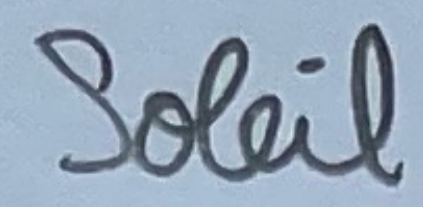}
         & soleil
         & \includegraphics[height=25mm]{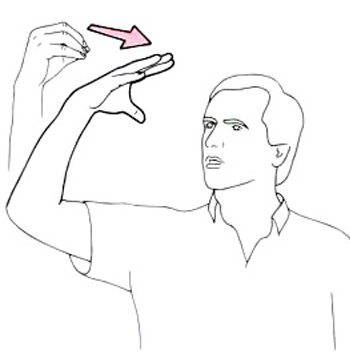}
    \end{tabular}
    \caption{Exemples de représentations graphiques spontanées}
    \label{fig:VD-régul-atom}
\end{figure}

On trouve également, dans les VD collectés, un certain nombre de choix et de dispositions graphiques récurrents, portant un sens identifiable et dont la réalisation est souvent la même, parfois au point où on peut parler de ``lecture associée''.
Il s'agit parfois de pictogrammes conventionnels ou icônes/symboles représentant des signes fixes, comme les exemples donnés en figure~\ref{fig:VD-régul-atom}.

\begin{figure}
    \centering
    \ref{enum:égal-FC}.~\includegraphics[height=15mm]{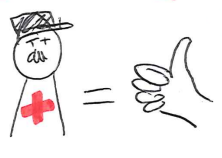}\hspace{10mm}
    \ref{enum:égal-lion}.~\includegraphics[height=15mm]{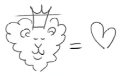}\hspace{10mm}
    \ref{enum:égal-vie}.~\includegraphics[height=12mm]{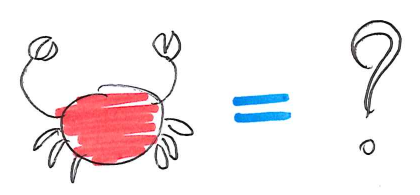}
    \caption{Extraits de VD comportant un signe ``égal''}
    \label{fig:VD-régul-egal}
\end{figure}

D'autres fois, il s'agit d'arrangements graphiquement plus complexes, combinant plusieurs sous-parties du dessin.
Par exemple, un signe d'égalité (``='') séparant une partie gauche $G$ d'une partie droite $D$ signifie en général que $D$ est une information sur $G$.
Des extraits collectés sont donnés en figure~\ref{fig:VD-régul-egal}, et leur interprétations en contexte sont :
\begin{enumerate}
    \item \label{enum:égal-FC} ``L'état de santé de Fidel Castro est bon.'' (F.~C. va bien.)
    \item \label{enum:égal-lion} ``Le lion est gentil.''
    \item \label{enum:égal-vie} ``Le cancer [c']est quoi ?'' (Qu'est-ce que le cancer ?)
\end{enumerate}

\begin{figure}
    \centering
    \includegraphics[width=.32\columnwidth]{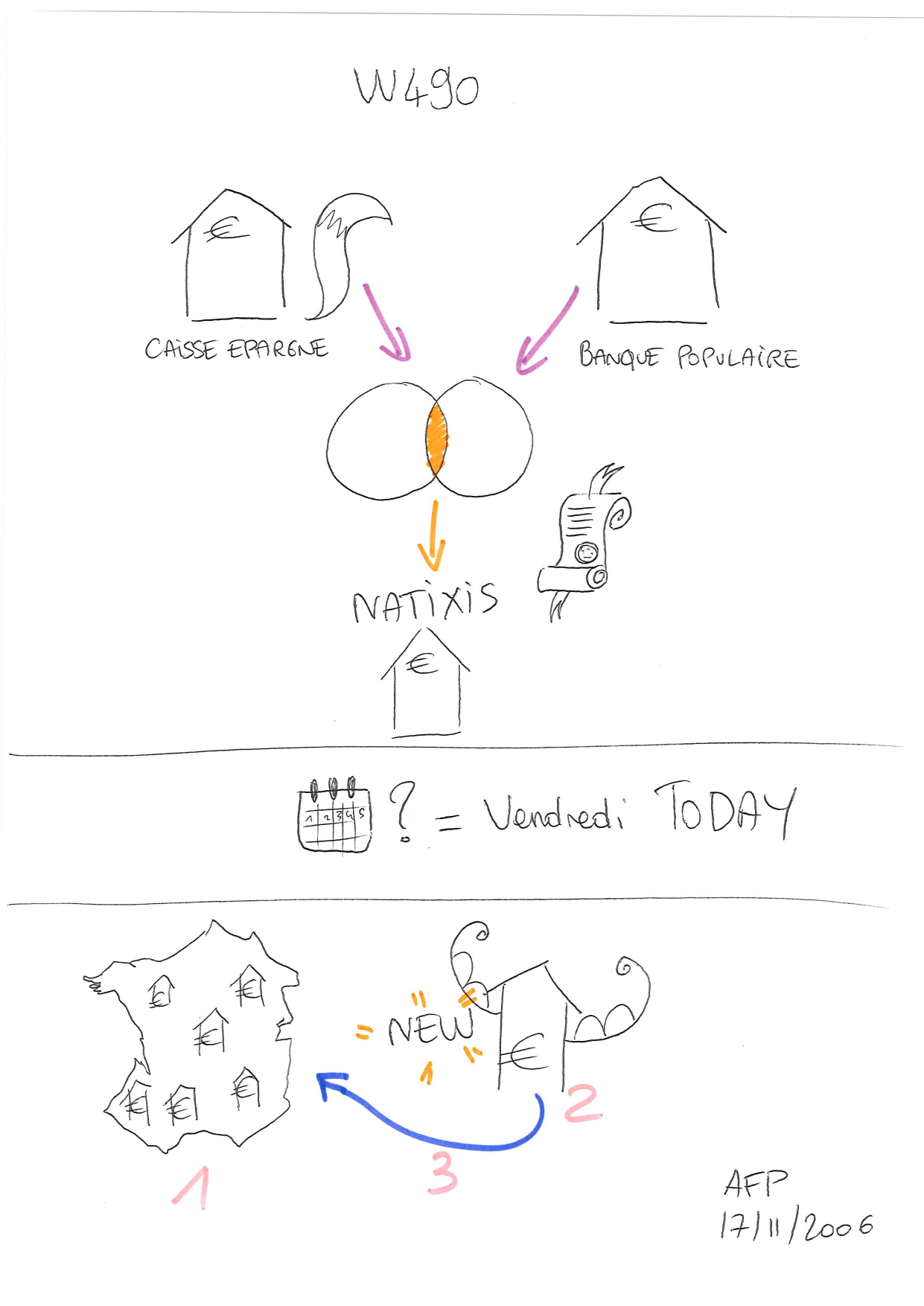}\hfill
    \includegraphics[width=.32\columnwidth]{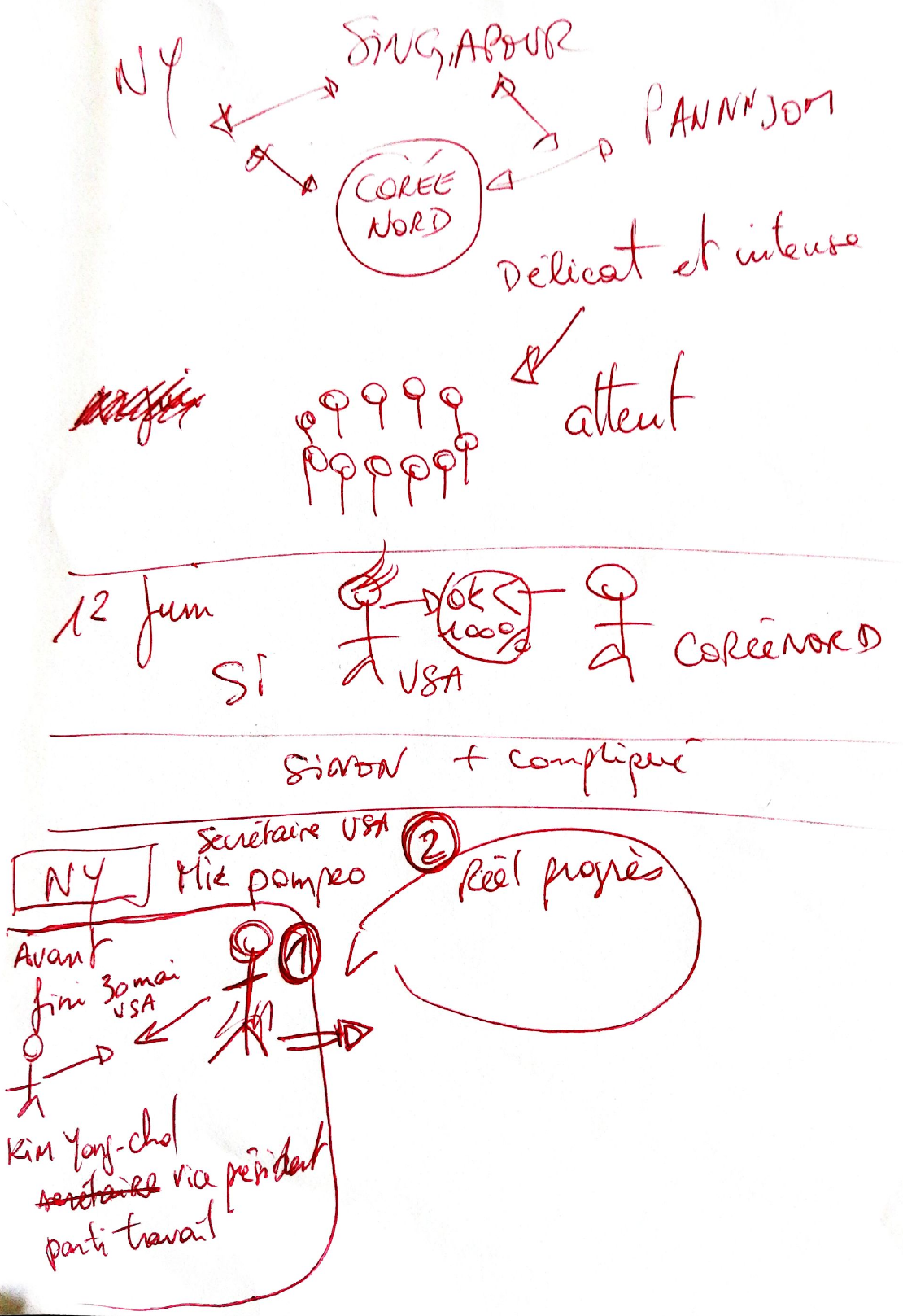}\hfill
    \includegraphics[width=.32\columnwidth]{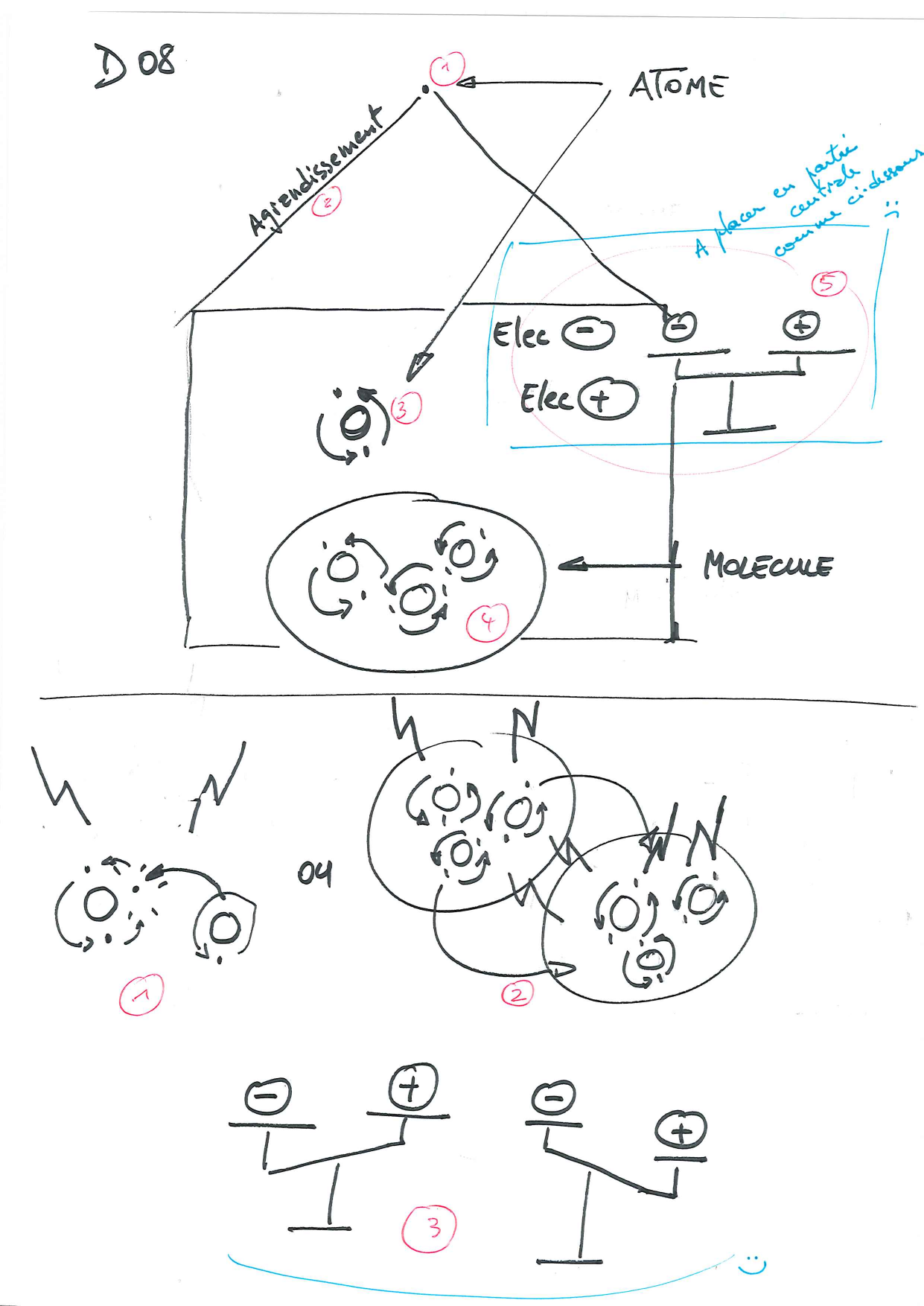}
    \caption{Pages de VD comportant des barres de contexte}
    \label{fig:VD-régul-contexte}
\end{figure}

Autre exemple, un VD comportant une barre séparant une partie haute $H$ d'une partie basse $B$ est toujours interprété comme la donnée d'un contexte $H$ pour $B$.
Cela inclut les cas de séquences temporelles, $B$ se déroulant finalement dans le contexte de $H$ étant fini avant lui.
Des exemples sont donnés en figure~\ref{fig:VD-régul-contexte}.

\begin{figure}
    \centering
    \begin{tabular}{cc}
        \includegraphics[height=10mm]{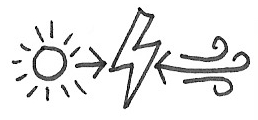}
         & \includegraphics[height=18mm]{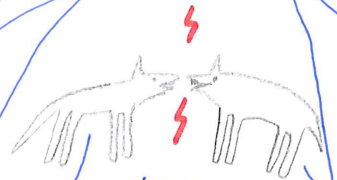}
        \\[5mm]
        \includegraphics[height=25mm]{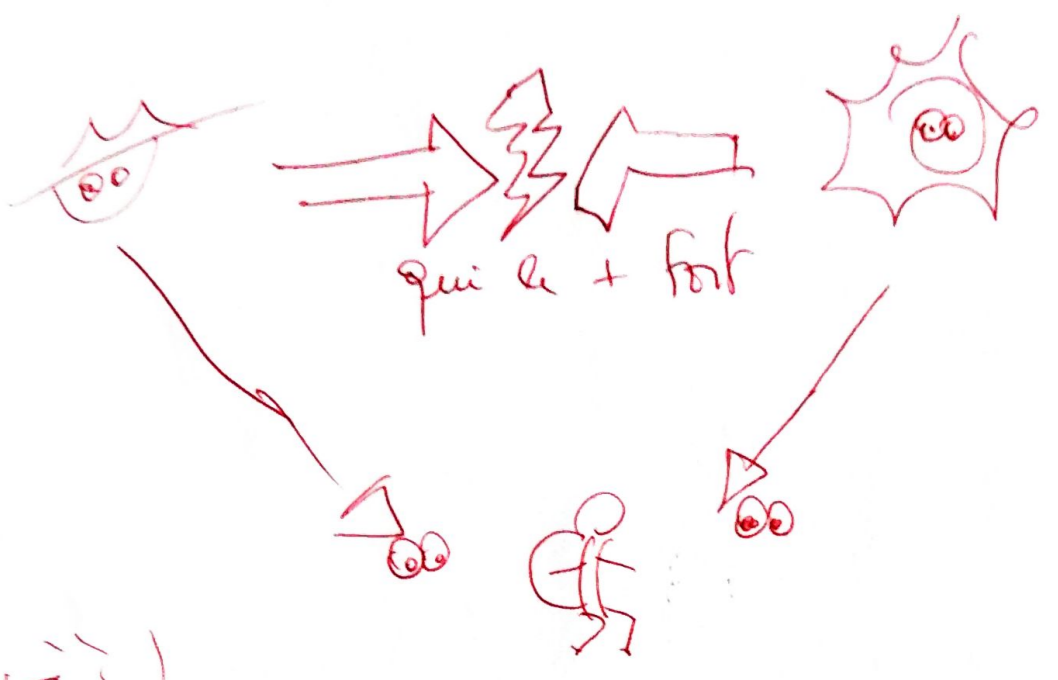}
         & \includegraphics[height=20mm]{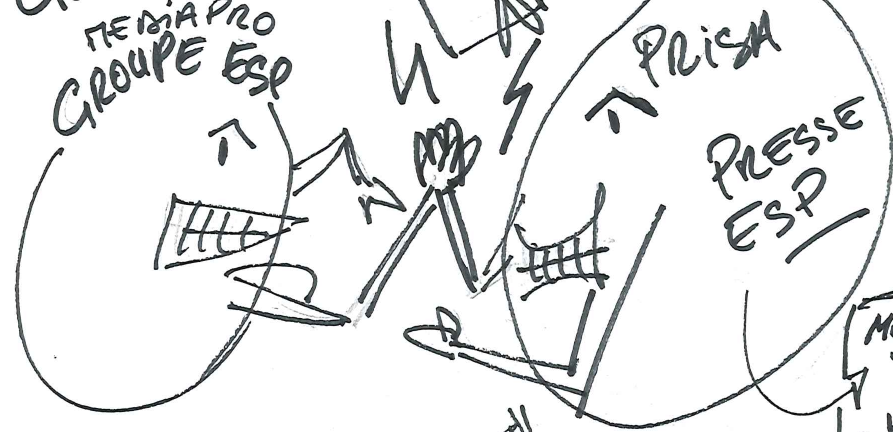}
    \end{tabular}
    \caption{Extraits de VD comportant des symboles ``éclair''}
    \label{fig:VD-régul-éclair}
\end{figure}

Ou encore, la séparation par un éclair signifiant une opposition entre deux entités, signées à la réalisation chacune d'un côté de l'espace.
Des exemples sont donnés figure~\ref{fig:VD-régul-éclair}.

Ainsi des régularités apparaissent malgré l'absence de système a priori.
Et en tant que pratique adoptée par certains, et par nature adaptée aux LS mais évoluant hors de tout système, les VD apparaissent comme une sorte d'intermédiaire entre :
\begin{itemize}
    \item les sytèmes réguliers adoptés, mais non-LS, p.~ex.\ le français écrit;
    \item et les systèmes réguliers conçus pour les LS, mais non-adoptés, p.~ex.\ SignWriting.
\end{itemize}

En les positionnant de la sorte, une piste apparaît pour obtenir un système qui soit à la fois régulier, adapté aux LS et adoptables pour les locuteurs, que nous avons choisi d'explorer : standardiser les VD.

\section{Standardisation des VD}

Standardiser les VD signifie pour nous créer un système graphique régulier et cohérent dont chaque production détermine sa lecture sans ambiguïté.
Un bon critère de test de cette faculté est par exemple la possibilité d'en générer automatiquement un rendu par un avatar signant.

Le problème est le risque que chaque contrainte de régularité apportée à la pratique des VD s'accompagne d'une perte d'adoptabilité.
En effet la pratique actuelle tient sans doute en partie à la liberté ressentie par les scripteurs.
Mais la présence de régularités spontanées déjà observées nous permet de supposer que celles-ci ne sont pas la source principale du rejet lorsqu'il existe.
Ainsi, nous faisons l'hypothèse déjà proposée \cite{SLS-2020-writing} qu'une standardisation qui respecterait les principes réguliers spontanément observés dans les VD serait d'autant plus facilement adoptable qu'elle serait inspirée de pratiques déjà habituelles.

Selon cette proposition, la liste des dispositions graphiques régulières présentées en section précédente devrait donc faire partie du standard, chacune imposant la lecture qui lui est associée.
Cela dit, leur nombre reste limité et on comprend que cette approche utilisée seule se heurtera nécessairement au problème de la couverture de la langue.
On peut envisager d'augmenter le système graphique avec des règles supplémentaires.
Notamment, bien que peu nombreuses, quelques recherches se sont intéressées aux parallèles existant entre les schémas pédagogiques et l'organisation de l'espace de signation en LS \cite{gui07}.
Cela dit, aucune n'a encore prétendu couvrir l'ensemble des phénomènes de la langue, ni être prescriptive au sens où la lecture pourrait être déterminée par les schémas.

Il a en revanche été constaté une proximité sémantique de plusieurs régularités VD avec des règles connues en \emph{AZee}, un système formel de représentation des énoncés en LS \cite{challant2022-corpusAZee}.
Initialement prévu pour la synthèse par avatar, ce système va prendre une place centrale pour compléter notre proposition car il a récemment démontré sa capacité couvrante.
Aussi nous en présentons un résumé en section suivante.

\section{AZee}

\begin{figure}
    \centering
    \includegraphics[height=30mm]{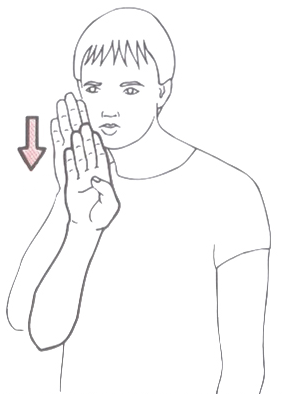}
    \caption{Forme de \AZee{gentil} en LSF}
    \label{fig:gentil}
\end{figure}

AZee est une approche formelle de représentation des énoncés signés basée sur l'identification de correspondances de sens (interprété) à forme (articulée) systématiques dans la langue considérée.
Sans supposition quant à l'existence ou la distinction de niveaux (morphologie, lexique, syntaxe...) ni de catégorisation syntaxique (verbe, nom, adjectif...), on définit des \emph{règles de production}, chacune associant, à un sens identifié qui lui donne en général son nom, une forme spécifiée à réaliser pour l'exprimer.
L'identification des règles se fait grâce à une méthodologie de recherche sur corpus \cite{mnh18} pour garantir une reproductibilité et une validité empirique.
Si l'approche est bien déclinable pour n'importe quelle LS a priori, les règles de production définies ainsi le sont donc pour une langue donnée.
La spécification formelle de la règle se fait avec le langage natif AZee, non détaillé ici, qui permet notamment de décrire la synchronisation des différentes articulations mises en jeu (orientations des doigts, position de la main...) en une valeur de type \AZee{SCORE} qui recouvre en somme les productions signées.
Par exemple en LSF, on définit la règle ``\AZee{gentil}'' comme portant le sens éponyme et produisant la réalisation représentée en figure~\ref{fig:gentil}, cette réalisation étant stable en LSF.

\begin{figure}
    \centering
    \includegraphics[height=15mm]{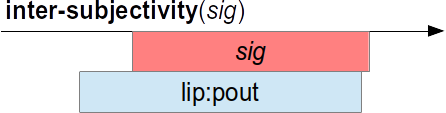}\hspace{5mm}
    ``lip:pout''~=~\includegraphics[height=10mm]{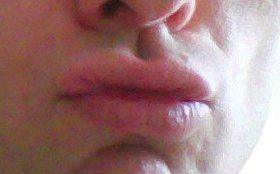}
    \caption{Arrangement temporel des formes pour \AZee{inter-subjectivity}}
    \label{fig:inter-subj}
\end{figure}

Ces règles peuvent être paramétrées lorsqu'une variation de forme est interprétable sémantiquement.
Par exemple, quelle que soit la production signée \emph{sig}, l'ajout d'un geste des lèvres synchronisé en avance sur \emph{sig} (fig.~\ref{fig:inter-subj}) est une forme signifiant un caractère consensuel à \emph{sig} : ``tout le monde est/serait d'accord que [\emph{sig}]''.
On définit donc la règle ``\AZee{inter-subjectivity}'', à un argument \emph{sig} de type \AZee{SCORE}, portant ce sens et produisant la synchronisation de la fig.~\ref{fig:inter-subj}, sens et forme dépendant ici des sens et forme de \emph{sig}.
L'ensemble des règles de production 

Combiner ces différentes règles de production, dont l'ensemble est appelé \emph{ensemble de production}, permet de construire des expressions AZee particulières, appelées \emph{expressions de discours}, générant des discours complets en LSF.
Par exemple, l'expression ci-dessous produit mécaniquement la réalisation signée pouvant être interprétée comme : ``généralement considéré comme gentil''.
Le signe ``\AZee{:}'' note l'\emph{application} de la règle dont les arguments sont indentés en-dessous, chacun sous une ligne spécifiant son nom.

\begin{minipage}[t]{\columnwidth}
\begin{verbatim}
:inter-subjectivity
  'sig
  :gentil
\end{verbatim}
\end{minipage}
\medskip

\begin{figure}
    \centering
    \includegraphics[height=15mm]{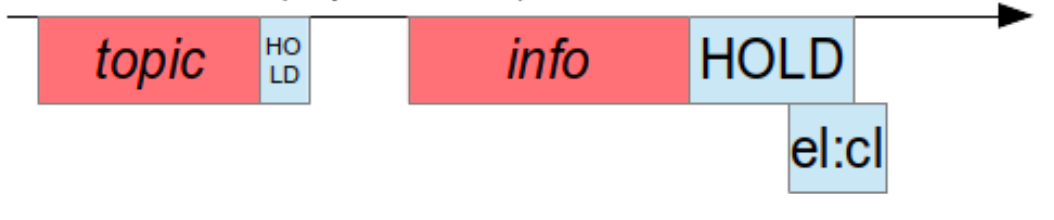}
    \caption{Arrangement temporel des formes pour \AZee{info-about}}
    \label{fig:info-about}
\end{figure}

\begin{figure}
    \centering
    \includegraphics[height=20mm]{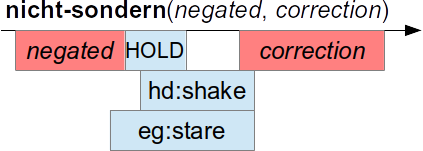}
    \caption{Arrangement temporel des formes pour \AZee{nicht-sondern}}
    \label{fig:nicht-sondern}
\end{figure}

De même, ``\AZee{info-about}(\emph{topic}, \emph{info})'' produit l'arrangement temporel de la figure~\ref{fig:info-about} et signifie ``\emph{info} à propos de \emph{topic}''.
Ou encore, ``\AZee{nicht-sondern}(\emph{nicht}, \emph{sondern})'' celui de la figure~\ref{fig:nicht-sondern} et signifie ``\emph{sondern}, et non pas \emph{nicht}''.
En supposant les deux règles éponymes simples (sans argument) \AZee{lion} et \AZee{méchant}, on peut donc construire l'expression ci-dessous signifiant ``le lion est gentil, et non méchant'', et générant le discours approprié en LSF.

\begin{minipage}[t]{\columnwidth}
\begin{verbatim}
:info-about
  'topic
  :lion
  'info
  :nicht-sondern
    'nicht
    :méchant
    'sondern
    :gentil
\end{verbatim}
\end{minipage}
\medskip

On peut ainsi construire des expressions de discours de plus en plus élaborées.
Un point important est que ces expressions permettent une lecture presque directe du sens global qu'elles représentent, tout en en déterminant les formes résultantes.

AZee s'est progressivement développé depuis plus d'une décennie et fait toujours l'objet de travaux pratiques pour en affiner l'ensemble de production de la LSF et le confronter à des corpus réels, et théoriques pour se positionner vis-à-vis des différentes grammaires formelles proposées en TAL.
Ces dernières années cependant, il a fait ses preuves sur la LSF en produisant d'une part un corpus d'expressions couvrant 1~h d'énoncés journalistiques d'une part \cite{challant2022-corpusAZee}, à savoir ceux constituant le corpus des \emph{40~brèves}, téléchargeables en ligne\footnote{\url{https://www.ortolang.fr/market/corpora/40-breves}}.
D'autre part, le modèle a aussi permis l'animation de l'avatar Paula \cite{Paula-SLTAT,Paula-lifelike}, notamment plusieurs phénomènes de langues restés inaccessibles avant AZee comme les constructions géométriques dans l'espace de signation continu \cite{McDonald-Filhol-2021-MT}.

\section{AZVD}

AZee est une approche orientée ordinateur et non humain, une expression en AZee ressemblant plus à du code informatique qu'à un schéma graphiquement organisé.
Mais son intérêt pour nous est double :
\begin{itemize}
    \item il s'agit d'un système formel, régulier et synthétisable, autant de buts recherchés pour notre standard graphique ;
    \item il a démontré une capacité de couverture de la langue, propriété qui semblait inatteignable en partant des seules régularités VD pour construire un standard.
\end{itemize}

Pour s'appuyer sur ces propriétés, nous proposons que, quel que soit le système graphique standard créé :
\begin{itemize}
    \item toute production graphique détermine une expression AZee unique de telle sorte qu'elle soit nécessairement synthétisable -- en termes formels ceci revient à définir une \emph{fonction} vers l'ensemble des expressions AZee possibles ;
    \item toute expression AZee puisse trouver au moins un antécédent graphique de telle sorte que le standard graphique puisse être lui-même réputé aussi couvrant qu'AZee -- nous visons donc que la fonction en question soit \emph{surjective}.
\end{itemize}

Considérons d'abord les dispositions graphiques régulières observées plus haut dans les VD.
Si l'on suppose qu'à chacune est associée une forme en LS déterminée avec un sens identifié d'une part, et qu'on dispose d'un ensemble de production AZee effectivement couvrant d'autre part, on peut alors trouver un patron d'expression AZee, ou \emph{template}, éventuellement paramétré par des éléments du dessin, qui la représente.
Dans ce cas, notre proposition est de spécifier une disposition graphique, ou \emph{layout}, générant le template AZee en utilisant ses éléments imbriqués pour en instancier les arguments.
Un tel layout peut faire usage d'icônes, spécifier un alignement ou des positions relatives entre ceux-ci, ainsi qu'y appliquer toute transformation telle qu'une homothétie ou une rotation.

Le cas trivial est celui du symbole atomique auquel on associe directement un template AZee fixe, sans argument variable.
Par exemple, suivant la régularité présentée figure~\ref{fig:VD-régul-atom}, on peut associer au symbole ``\includegraphics[height=2ex]{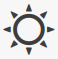}'' l'expression AZee:

\begin{minipage}[t]{\columnwidth}
\begin{verbatim}
:soleil
\end{verbatim}
\end{minipage}
\medskip

\begin{figure}
    \centering
    (a)~\includegraphics[height=18mm]{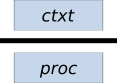}\hspace{5mm}
    (b)~\includegraphics[height=10mm]{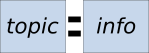}\hspace{5mm}
    (c)~\includegraphics[height=14mm]{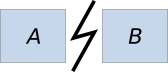}
    \caption{Layouts AZVD inspirés des régularités VD}
    \label{fig:AZVD-vers-cible}
\end{figure}

Les régularités impliquant des parties variables peuvent être encodées graphiquement avec la même récursivité pour leur contenu. 
Par exemple, les ``barres de contexte'' régulières observées figure~\ref{fig:VD-régul-contexte} incitent à définir la disposition de la figure~\ref{fig:AZVD-vers-cible}.a, comportant deux parties variables \emph{ctxt} et \emph{proc} respectivement au-dessus et en-dessous de la barre horizontale, et à y associer la lecture systématique donnée par le template ci-dessous, dont les parties entre crochets correspondant directement aux sous-parties du même nom dans le layout :

\begin{minipage}[t]{\columnwidth}
\begin{verbatim}
:context
  'ctxt
  [ctxt]
  'proc
  [proc]
\end{verbatim}
\end{minipage}
\medskip

De même, en vertu de la régularité figure~\ref{fig:VD-régul-egal}, la disposition fig.~\ref{fig:AZVD-vers-cible}.b trouve une correspondance directe avec la règle \AZee{info-about} et ses arguments \emph{topic} et \emph{info}.

Parfois des templates AZee plus complexes sont nécessaires, le principe restant toutefois le même.
Par exemple, la disposition de la figure~\ref{fig:AZVD-vers-cible}.c, inspirée de la régularité présentée figure~\ref{fig:VD-régul-éclair}, génère le template ci-dessous, dont la forme résultante est le placement successif de \emph{A} d'un côté puis \emph{B} de l'autre avec une tenue et un clignement des yeux après chaque (ceci généré par la règle \AZee{each-of}), et qui signifie une opposition entre les deux :

\begin{minipage}[t]{\columnwidth}
\begin{verbatim}
:each-of
  'items
  list
    :about-point
      'pt
      ^Lssp
      'locsig
      [A]
    :about-point
      'pt
      ^Rssp
      'locsig
      [B]
\end{verbatim}
\end{minipage}
\medskip

Notons que parfois, plusieurs dispositions VD existent produisant le même template AZee de manière régulière.
Par exemple, les barres de contexte peuvent être observées avec une disposition horizontale ou verticale.
Ceci ne pose pas de problème ; plusieurs \emph{variants} peuvent être définis pour une même correspondance AZee.

\begin{figure}
    \centering
    \includegraphics[height=15mm]{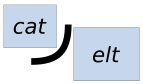}
    \caption{Layout défini pour \AZee{category}(\emph{cat}, \emph{elt})}
    \label{fig:AZVD-category}
\end{figure}

Cette méthode assure des dispositions VD dont à la fois la lecture et le dessin sont réguliers.
Mais nous l'avons dit, celles-ci seules ne suffiront pas à constituer la surjection nécessaire à couvrir l'ensemble des expressions AZee possibles, la plupart restant ici sans antécédent graphique.
Une manière triviale de garantir la fonction surjective posée au départ comme objectif est d'augmenter l'inventaire des layouts disponibles en en fabriquant un par règle de production n'ayant pas déjà d'antécédent direct par la méthode précédente.
Par exemple, la règle de production \AZee{category} peut se voir attribuée le layout figure~\ref{fig:AZVD-category}, dont les parties notées \emph{cat} et \emph{elt} correspondent aux deux arguments de la règle.
De même que ceux définis à partir des VD observés, on peut aussi proposer différents variants pour une même règle AZee.

Avec cette méthode, les layouts créés ne sont pas inspirés de pratiques régulières observées, et pourraient par conséquent être moins susceptibles d'être adoptés.
De plus, les règles de production AZee qu'ils encodent peuvent elles-mêmes être plus ou moins bien connues ou appréhendées par les locuteurs.
Cela dit, celles-ci représentent en principe bien des opérations sémantiques de la langue empiriquement mises en évidence, donc qu'elles soient consciemment ou non intériorisées par les locuteurs, on peut supposer qu'elles ont au moins un sens accessible et utile à la ``pensée signe''.
Pour maximiser l'adoptabilité en pratique de notre script, nous proposons en outre pour chaque layout ainsi défini qu'il soit également :
\begin{itemize}
    \item facile à retenir, en le rendant iconique de ce qu'il représente quand c'est possible ;
    \item préférentiellement représentatif du sens (logographique), sauf rapport trop faible entre facilité de représentation du sens et saillance de la forme -- ce basculement phonographique a en effet été observé dans les pratiques spontanées ;
    \item ait au moins un variant facile à dessiner, donc monochrome, sans zone à remplir au crayon et avec faible nombre de traits.
\end{itemize}

\begin{figure}
    \centering
    \includegraphics[width=.9\columnwidth]{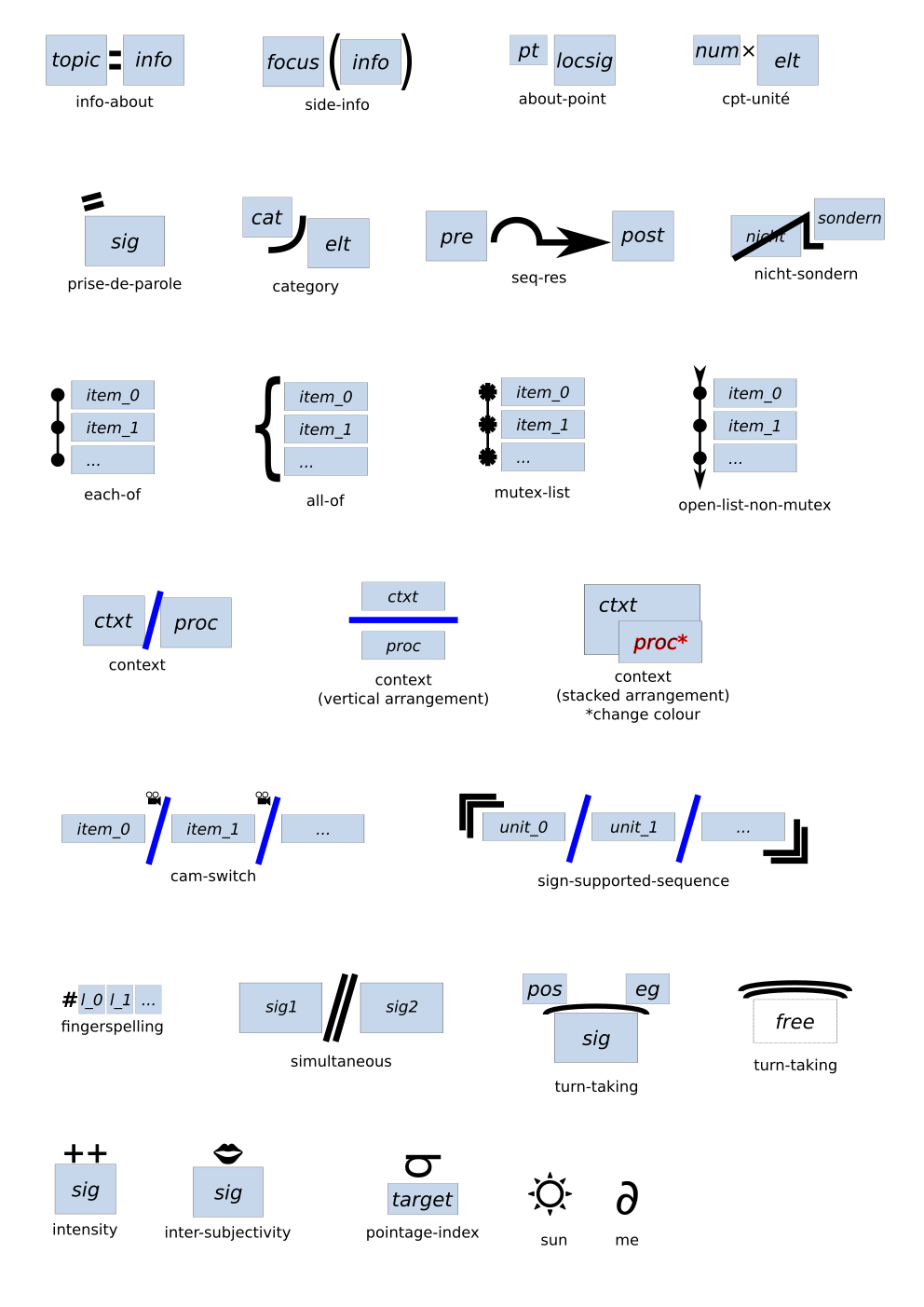}
    \caption{Planche des layouts AZVD de production AZee}
    \label{fig:AZVD-planche}
\end{figure}

À titre indicatif, nous donnons une planche en figure~\ref{fig:AZVD-planche} avec des layouts proposés pour les principales règles de production AZee.

\begin{figure}
    \centering
    \includegraphics[width=60mm]{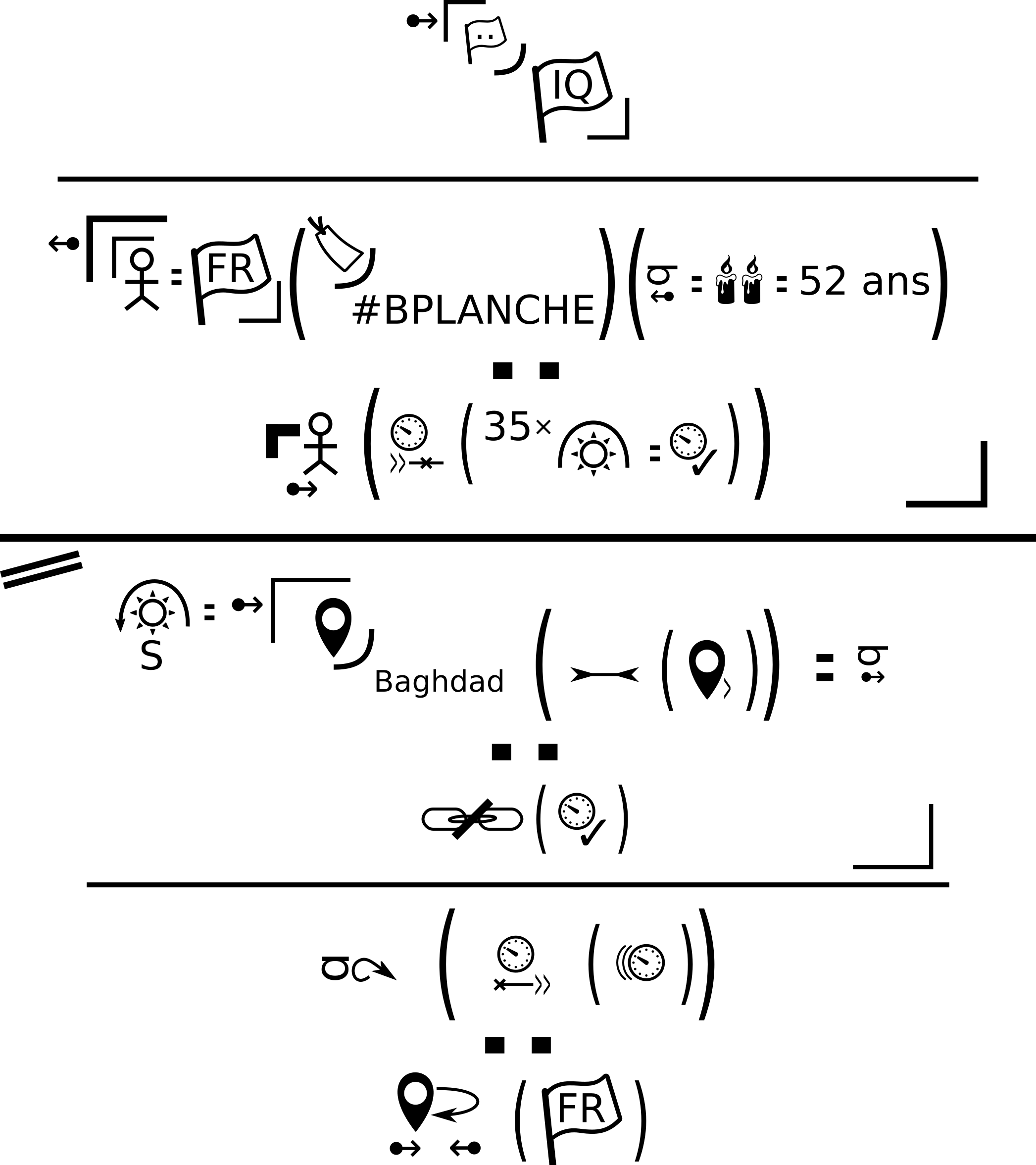}
    \caption{AZVD pour une brève journalistique complète}
    \label{fig:1R-JP}
\end{figure}

Ces deux méthodes garantissent conjointement une fonction surjective vers l'ensemble des expressions AZee possibles, avec un ensemble de définition composés d'éléments au plus proche de ce que l'on peut envisager adoptable.
L'ensemble des layouts ainsi formé et la possibilité de les combiner entre eux pour construire des schémas planaires (2D) arbitrairement complexes est ce que nous nommons le script \emph{AZVD}, pour \emph{AZee verbalising diagram}.
La figure~\ref{fig:1R-JP} est un exemple de schéma AZVD complet représentant l'entrée ``1R-JP'' du corpus des \emph{40~brèves} cité plus haut, dont la production en LS dure 20~s.

\section{Éditeur AZVD}

Nous avons décidé de faire un éditeur de schémas AZVD capable de générer les expressions AZee correspondantes pour tester d'abord, puis tenter ensuite de disséminer notre proposition.

\subsection{Spécification et rendu des layouts}

Nous avons défini précédemment que les layouts étaient par nature récursifs, et qu'ils sont susceptibles de mettre en jeu des transformations géométriques.

Pour les spécifier, nous avons construit le système de construction de layout suivant.
Il faut préciser, pour chaque layout, les élements qui le composent ainsi que leur organisation graphique.
Les layouts peuvent être de différentes nature :
\begin{itemize}
  \item une image ;
  \item du texte ;
  \item une \emph{zone à remplir} où est attendu un nouvel AZVD ;
  \item une liste d'éléments, chacun un nouvel AZVD ;
  \item ou plusieurs des éléments ci-dessus combinés.
\end{itemize}\par
\begin{figure}
    \centering
    \includegraphics[width=.45\columnwidth]{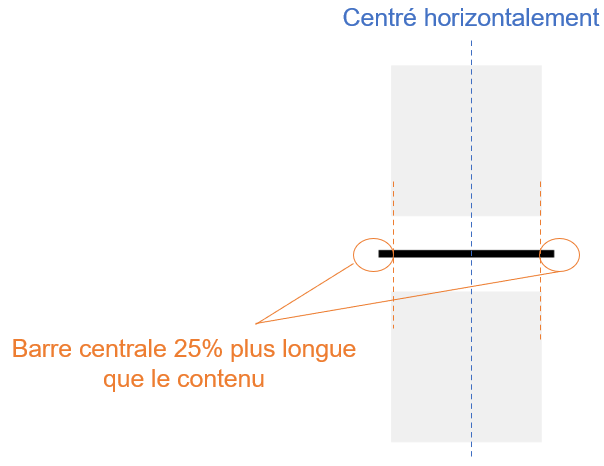}\hfill
    \includegraphics[width=.5\columnwidth]{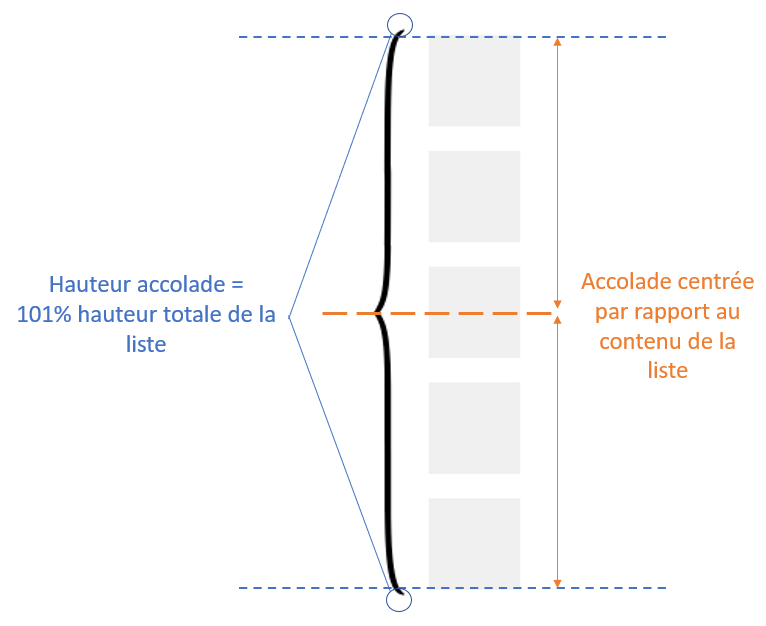}\\[5mm]
    \includegraphics[width=.8\columnwidth]{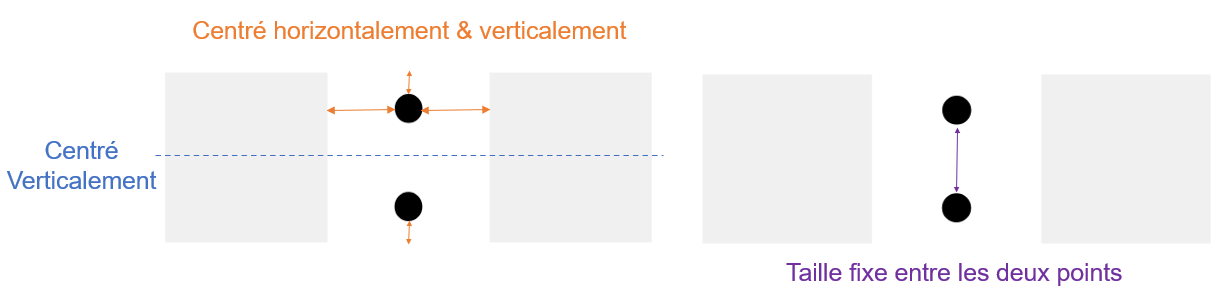}
    \caption{Spécifications de layouts -- en haut à gauche : cf.\ fig.~\ref{fig:AZVD-vers-cible}.a ; en haut à droite : layout créé pour la règle \AZee{all-of} (agrégat d'éléments, liste en argument) ; ligne du bas : cf.\ fig.~\ref{fig:AZVD-vers-cible}.b}
    \label{fig:spec-layouts}
\end{figure}

Pour ce dernier, la combinaison spécifie les relations spatiales et les relations de taille des éléments, comme illustré figure~\ref{fig:spec-layouts}.

\begin{figure}
    \centering
    \includegraphics[height=20mm]{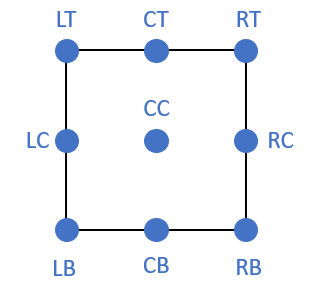}
    \caption{Points remarquables d'un élément graphique AZVD}
    \label{fig:alignPts}
\end{figure}

\begin{figure}
    \centering
    \includegraphics[width=.8\columnwidth]{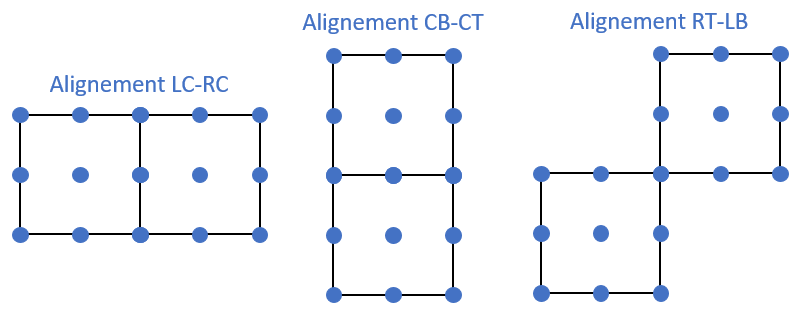}
    \caption{Exemples d'alignements, utilisés pour les layouts des figures \ref{fig:AZVD-vers-cible}.b (gauche) et \ref{fig:AZVD-vers-cible}.a (milieu), et pour la règle \AZee{nicht-sondern} (droite, cf.\ fig.~\ref{fig:AZVD-planche})}
    \label{fig:exAlignPts}
\end{figure}

Pour ces spécifications graphiques, nous avons implémenté : 
\begin{itemize}
  \item des points remarquables pour chaque élément graphique quel qu'il soit, cf.\ fig\ref{fig:alignPts} ;
  \item un système d'alignement de ces points entre eux, cf.\ fig.~\ref{fig:exAlignPts}, éventuellement avec des spécifications de décalage (translations) ;
  \item la possibilité de redimensionner des éléments l'un par rapport à l'autre ou par rapport à une échelle fixe.
\end{itemize}

Une fois spécifiés, pour en faire le rendu, nous avons choisi une représentation vectorielle des éléments graphiques (SVG dans notre cas).
Cette technologie donne une description formelle et récursive du contenu d'une image et permet d'appliquer des transformations (translation, rotation, homothétie) aux éléments contenus.

\begin{figure}
    \centering
    \includegraphics[width=.8\columnwidth]{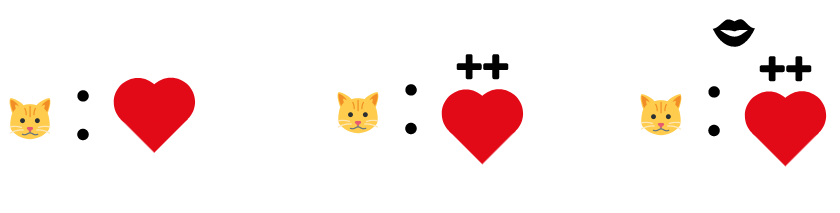}
    \caption{Constructions AZVD}
    \label{fig:constr-AZVD}
\end{figure}

Grâce à ce système associant la spécification par layouts et le rendu de dessin par SVG, les éléments s'alignent et se mettent en forme automatiquement lors de la construction d'un schéma dans l'éditeur, et ce quelle que soit leur complexité.
Nous présentons une progression simple d'exemples en figure~\ref{fig:constr-AZVD}.
Notons également que grâce aux spécifications de taille dans les layouts, il est possible de normaliser l'ensemble des éléments atomiques, leur donnant à chacun une taille nominale, et ainsi donner une cohérence visuelle à l'ensemble.

\subsection{Génération d'expressions AZee}

Au cahier des charges se trouve également la génération des expressions AZee résultant des combinaisons graphiques effectuées par l'utilisateur, chaque dessin déterminant une unique expression AZee.
Un identifiant est donné à chaque zone à remplir dans les layouts, et utilisé comme paramètre dans les templates associés.
Le contenu des zones à remplir est récursivement reporté le template AZee pour générer l'expression AZee complète.
Par exemple, les trois schémas de la figure~\ref{fig:constr-AZVD} génèrent les expressions suivantes, respectivement de gauche à droite :

\begin{minipage}{.3\columnwidth}
\begin{verbatim}
:info-about
  'topic
  :chat
  'info
  :gentil
\end{verbatim}
\end{minipage}
\hfill
\begin{minipage}{.3\columnwidth}
\begin{verbatim}
:info-about
  'topic
  :chat
  'info
  :intensity
    'sig
    :gentil
\end{verbatim}
\end{minipage}
\hfill
\begin{minipage}{.3\columnwidth}
\begin{verbatim}
:inter-subjectivity
  'sig
  :info-about
    'topic
    :chat
    'info
    :intensity
      'sig
      :gentil
\end{verbatim}
\end{minipage}

\subsection{Variants}

\begin{figure}
    \centering
    \includegraphics[frame,height=40mm]{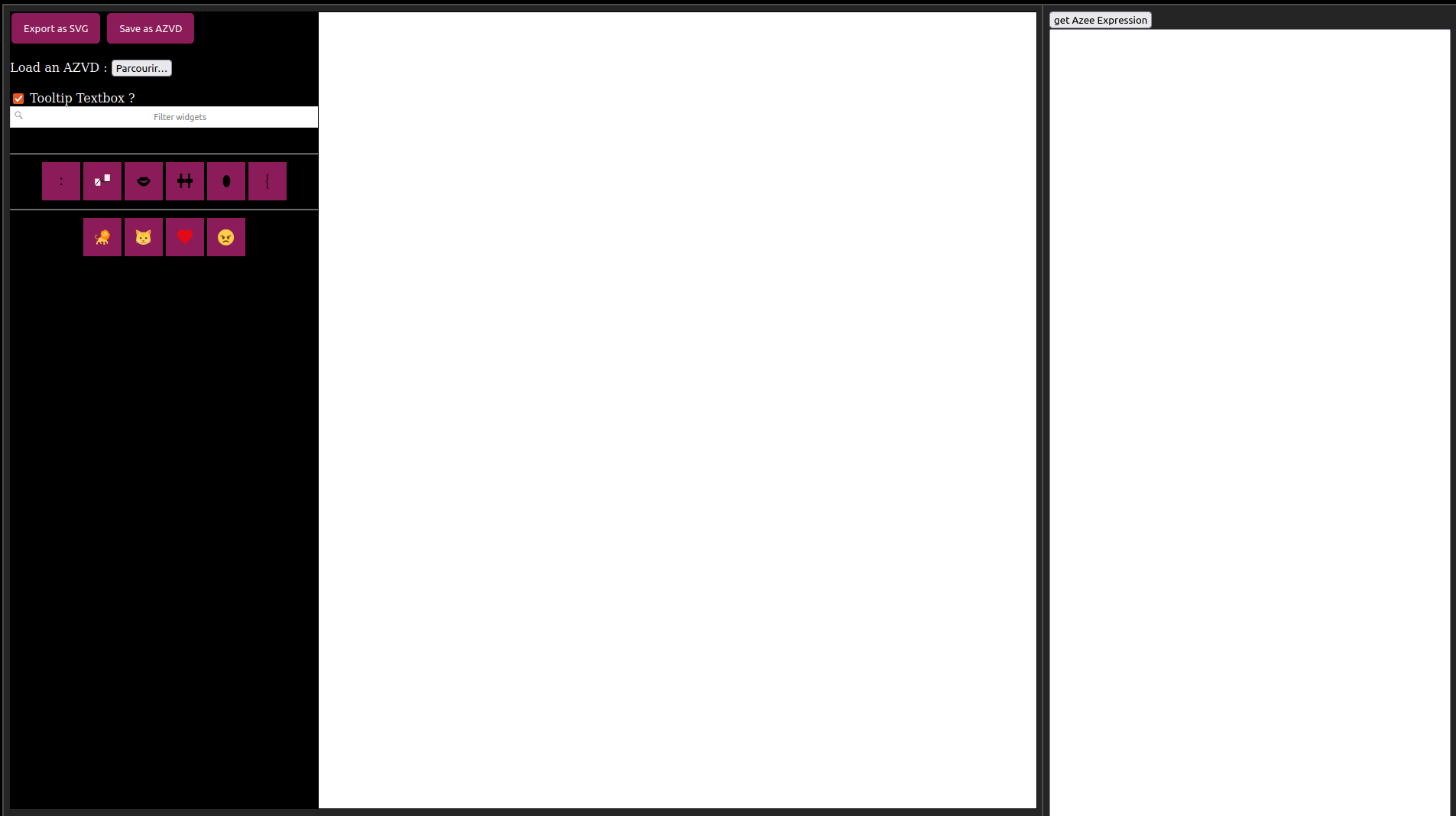}
    \caption{L'éditeur AZVD}
    \label{fig:editeurAZVD}
\end{figure}

\begin{figure}
    \centering
    \includegraphics[frame, height=20mm]{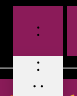}
    \caption{Menu de variants, ici incluant une version verticale de la fig.~\ref{fig:AZVD-vers-cible}.b}
    \label{fig:menuVariants}
\end{figure}

Dans l'interface de l'éditeur, la liste des éléments disponibles est visible dans la zone à gauche (cf.\ fig.~\ref{fig:editeurAZVD}).
Comme évoqué dans la section précédente, AZVD permet la définition de variants.
Dans l'éditeur, nous avons décidé de regrouper les variants produisant le même AZee dans un menu déroulant pour ne pas surcharger l'interface (cf.\ fig.~\ref{fig:menuVariants}).

\section{Perspectives}

Nous avons défini une représentation graphique pour la LSF, basée sur le modèle AZee.
Notre démarche a tenté de garantir la couverture de la langue d'AZee, et de maximiser l'adoptabilité du système, notamment en répercutant les régularités observées dans les pratiques de schématisation (VD) spontanées.
Cette section conclut en présentant quelques perspectives d'utilisation et de discussion, en particulier pour faire évoluer l'éditeur et le script lui-même dans le sens de leur adoptabilité, et pour caractériser et situer le script et ses règles vis-à-vis des autres, déjà adoptés.

\subsection{Tests et évolution du script}

Notre proposition complète compte deux parties, à savoir AZVD d'une part et l'éditeur d'autre part.
Les évaluer conjointement à travers une simple utilisation de l'éditeur rendrait difficile l'interprétation des résultats, l'expression d'un rejet de l'éditeur pouvant par exemple être attribué à son interface autant qu'aux fondements AZVD eux-mêmes.
Il convient donc de les distinguer.

Dans un premier temps, puisque AZVD comporte une correspondance directe pour chaque règle de production AZee, l'éditeur pourra être utilisé comme un éditeur graphique d'AZee, utile aux chercheurs travaillant avec ce modèle et à son développement.
Ceux-ci maîtrisant AZee et habitués au formalisme, ils pourront utiliser l'éditeur comme une simple assistance graphique à l'édition d'expressions.
Ceci fournit un moyen d'évaluer cet éditeur, son interface et sa capacité à générer des expressions AZee, avec un biais du modèle sous-jacent limité.
On pourra aussi mesurer cette assistance par la réduction du temps nécessaire à la réalisation des expressions.

Dans un deuxième temps, et les problèmes majeurs d'interface réglés, nous souhaitons mettre AZVD entre les mains d'usagers de la langue pratiquant déjà le VD, c'est-à-dire non-connaisseurs d'AZee mais pour qui la démarche de schématisation est déjà naturelle.
Si cette démarche relève en partie d'une évaluation de la proposition AZVD, elle a en réalité plutôt vocation à associer le public intéressé à un processus itératif pour l'élaboration du script pour le faire évoluer, par exemple en intégrant de nouveaux layouts souhaités, en abandonnant des variants inutiles...
Une perspective motivante est la réutilisation d'AZVD par les utilisateurs, au crayon hors cadre de l'éditeur.
Leurs manières d'abuser du script et de déroger aux règles posées a priori seront autant de pistes nouvelles pour les faire évoluer dans le sens d'un script plus adoptable.

\subsection{Linéarisation}

Intéressons-nous au script lui-même et à la possibilité pour les schémas AZVD de s'étendre sur une surface arbitrairement grande par imbrication récursive des layouts qui le composent.
Cette propriété du script d'être ainsi planaire (2D) et récursif contraste radicalement avec les scripts adoptés pour écrire les langue non signées.
En effet, ceux-ci sont en premier lieu linéaires, autrement dit représentent toujours une séquence d'unités au premier niveau d'organisation.

La composition planaire récursive a bien été régulièrement observée dans les VD étudiés, donc ne s'oppose pas en soi à l'adoptabilité du script.
Aussi, d'autres scripts non linguistiques comme la notation mathématique épousent la même propriété, et sont très largement adoptés.
Mais il est certain qu'en vue de représenter des discours longs, une telle propriété pose un problème pratique et une certaine linéarisation s'impose vu la limite nécessairement posée par les supports (écran ou page papier).
En outre, il est arrivé au cours de nos recherches que soit envisagé le sous-titrage comme utilisation du script.
Le sous-titre traditionnel, en français par exemple, découpe le texte en unités de durées limitées, et leur affichage est déterminé par le temps réel utilisé par la vidéo.
De même, il conviendrait de pouvoir séquencer un AZVD global en unités plus petites unités, chacune pouvant être affichée séparément mais sans que la recomposition du sens global soit perdue.

Suite à de premiers essais inspirés par les pauses entre sous-titres à l'écrit, notre conjecture est que les segmentations les plus naturelles tombent toujours dans l'intervalle de temps séparant le \emph{ctxt} et le \emph{proc} d'une règle \AZee{context}.
Cela dit, il est possible que certaines applications de \AZee{context} ne se prêtent pas à une séparation de leur contenu en deux schémas successifs, par exemple lorsqu'ils sont imbriqués dans certains règles exigeant une vue simultanée de leurs arguments.
Peut-être sera-t-il également nécessaire de couper les AZVD à d'autres endroits.
Davantage de recherche est nécessaire pour répondre, mais la perspective de linéariser le script pour le rendre utilisable sur de longs discours tout en restant lisible et interprétable est tout à fait motivante et pourrait à long terme en faire un candidat à l'écriture.

\subsection{De la surjection à l'injection, et parallèles avec l'écriture}

Nous avons construit AZVD de sorte à pouvoir définir une fonction surjective vers l'ensemble des expressions AZee possibles afin de garantir une couverture de la langue cible.
Cela dit, il est tout à fait possible que cet ensemble soit en réalité lui-même un ensemble surdimensionné, et que toutes les expressions AZee théoriquement possibles ne soient pas nécessaires pour couvrir la LSF.
Des recherches sont d'ailleurs en cours sur les contraintes s'appliquant aux expressions AZee, et sur la définition grâce à celles-ci de grammaires formelles pour les LS.
La discussion est hors du champ de cet article, mais il pourrait à terme être établi formellement que certaines expressions AZee sont linguistiquement interdites (grammaticalement inacceptables), ou certaines règles jamais utilisées hors de certains patrons d'expressions en nombre fini...
Selon les cas, il pourrait alors être justifié de laisser des expressions AZee sans antécédents graphiques et abandonner la propriété surjective de AZVD dans AZee, posée comme objectif méthodologique de départ.

Inversement, on peut constater que la fonction construite n'est pas une injection.
En effet, l'existence de variants mais aussi de templates complexes pouvant se ramener à une imbrication de templates plus simples, plusieurs antécédents graphiques ont ici la même image en AZee, et donc la même forme déterminée à la lecture.
Il serait intéressant de se demander si à terme, l'existence de multiples options pour représenter la même lecture est appréciée ou si au contraire il serait préférable de contraindre les choix d'antécédents lorsqu'il existent, et ainsi se rapprocher voire imposer un comportement injectif.
De même, la discussion sort du cadre de cet article mais elle est une perspective intéressante, notamment si elle se donne des parallèles avec les scripts existants.
En effet, ne serait-ce qu'en français, on trouve des cas de variantes orthographiques qui semblent être au choix du scripteur et sans incidence sur la lecture (\emph{clé}/\emph{clef}, \emph{cuillère}/\emph{cuiller}...), mais aussi des variantes typographiques (a/\textscripta, \textg/\textscriptg...) ou de casse (majuscule/minuscule) pour les lettres.
Certains choix sont contraints par la grammaire, d'autres sont esthétiques...
Jusqu'où pourra-t-on retrouver ces notions et propriétés ou établir les parallèles dans un script tel qu'AZVD ?

\section*{Remerciements}
La recherche à l'origine de cet article a été en partie financée par le projet européen \emph{EASIER} (\emph{Intelligent Automatic Sign Language Translation}), accord de financement numéro~101016982 du programme cadre \emph{Horizon~2020}. Source des images pour les figures \ref{fig:VD-régul-atom} (à droite) et \ref{fig:gentil} : \cite{IVT-dico-1997}.

\bibliography{AZVD-article}
\bibliographystyle{plain}
\end{document}